\documentclass[letterpaper]{article} 
\usepackage{aaai2026}
\usepackage{times}  
\usepackage{helvet}  
\usepackage{courier}  
\usepackage[hyphens]{url}  
\usepackage{graphicx} 
\usepackage{natbib}  
\usepackage{caption} 
\usepackage{amssymb}
\usepackage{amsmath}
\usepackage{booktabs}
\usepackage{caption}
\usepackage{multirow}
\usepackage{tabularray}
\usepackage{subcaption}

\usepackage{algorithm}
\usepackage{algorithmic}

\usepackage{newfloat}
\usepackage{listings}
\DeclareCaptionStyle{ruled}{labelfont=normalfont,labelsep=colon,strut=off} 
\floatstyle{ruled}
\newfloat{listing}{tb}{lst}{}
\floatname{listing}{Listing}
\title{Amber Pruner: Leveraging N:M Activation Sparsity\\for Efficient Prefill in Large Language Models}
\author {
    Tai An\textsuperscript{\rm 1}\thanks{Corresponding author: antai2018@ia.ac.cn.},
    Ruwu Cai\textsuperscript{\rm 1},
    Yanzhe Zhang\textsuperscript{\rm 1},
    Yang Liu\textsuperscript{\rm 1},
    Hao Chen\textsuperscript{\rm 1},\\
    Pengcheng Xie\textsuperscript{\rm 1},
    Sheng Chang\textsuperscript{\rm 1},
    Yiwu Yao\textsuperscript{\rm 1},
    Gongyi Wang\textsuperscript{\rm 1}
}
\affiliations {
    \textsuperscript{\rm 1}Huawei Technologies Co., Ltd\\
    \small
    antai3@huawei.com, cairuwu@huawei.com, zhangyanzhe1@huawei.com,liuyang1051@huawei.com, chenhao241@huawei.com,\\xiepengcheng2@huawei.com,leo.chang@huawei.com, yaoyiwu@huawei.com, techw.wanggongyi@huawei.com
}

\usepackage{bibentry}

\begin{document}

\maketitle

\begin{abstract}
In the era of large language models (LLMs), N:M sparsity has emerged as a structured compression technique critical for accelerating inference. While prior work has primarily focused on weight sparsity, it often suffers from significant accuracy degradation. Activation sparsity, though promising, is typically training-dependent and faces challenges in generalization. To address these limitations, we introduce Amber Pruner, a training-free N:M activation sparsity method designed specifically for the prefill stage, targeting the acceleration of linear projection layers in LLMs. Extensive experiments across multiple models and sparsity ratios (2:4, 4:8, and 8:16) demonstrate that Amber Pruner can effectively sparsify and accelerate more than 55\% of linear computations without requiring model retraining. To further enhance generality and efficiency, we propose Outstanding-sparse, a unified framework that integrates Amber Pruner with post-training W8A8 quantization. Our approach preserves strong performance across a range of downstream tasks, with notable advantages in generative tasks. This work pioneers a new frontier in activation sparsity, providing foundational insights that are poised to guide the co-evolution of algorithms and architectures in the design of next-generation AI systems.
\end{abstract}


\section{Introduction}
The release of DeepSeek V3 \cite{liu2024deepseek} marks a significant milestone in the advancement of sparse computing, reaffirming its status as a crucial direction for next-generation AI systems. Among various sparsity techniques, N:M sparsity has garnered widespread attention from both academia and industry due to its structural compatibility with neural networks and strong alignment with mainstream hardware architectures. N:M sparsity enforces a constraint where only N non-zero elements are retained within every M consecutive elements. As a semi-structured sparsity pattern, it enables substantial inference acceleration on sparsity-aware general-purpose hardware such as GPUs and NPUs. Consequently, N:M sparsity is increasingly seen as a promising path toward the efficient deployment of large language models (LLMs).

The concept of N:M sparsity traces back to NVIDIA’s introduction of 2:4 sparsity in the Ampere architecture \cite{mishra2021accelerating}, which demonstrated improvements in compute throughput and memory bandwidth through semi-structured weight pruning. In the context of LLMs, training-free methods such as SparseGPT \cite{frantar2023sparsegpt} and Wanda \cite{sun2023simple} have emerged as representative approaches for applying N:M constraints to model weights. These methods evaluate the importance of weight parameters and generate pruning masks accordingly. However, such techniques often suffer from substantial accuracy degradation on challenging benchmarks like MMLU \cite{hendryckstest2021}—where performance drops can exceed 20\%, limiting their practicality for real-world deployment.

Instead of focusing on weight sparsity, another line of research centers on activation sparsity. Works like Q-Sparse \cite{wang2024q} and TEAL \cite{liu2024training} apply dynamic sparsification to activation tensors during the decoding phase to reduce memory access costs. These methods yield notable latency reductions under single-batch inference. However, in multi-batch scenarios, the bottleneck limits the effectiveness of such techniques at scale. Other efforts aim to increase activation sparsity by modifying activation functions like Squared-ReLU \cite{wang2024q, haziza2025accelerating}. While these training-aware methods can often recover full model performance across downstream tasks, they raise concerns regarding generalization and semantics.

To address these challenges, we propose Amber Pruner, a training-free algorithm based on N:M activation sparsity. Noting that the linear projection layers during the prefill phase in LLM inference exhibit particularly high compute density, Amber Pruner employs a top-k strategy to determine N:M input activation masks within these layers. It further integrates a Robust-Norm Scoring mechanism and avoids pruning modules known to be sensitive (e.g., o\_proj and up\_proj), reducing the risk of accuracy degradation. Under 8:16 sparsity, Amber Pruner accelerates over 55\% of computations in linear projection while maintaining strong performance: Zero-shot tasks incur less than 1\% average accuracy loss, and the model’s generation ability remains unaffected, validating the practicality of N:M sparsity for LLMs.

Notably, Amber Pruner is orthogonal and highly compatible with existing quantization techniques. We introduce a synergistic optimization strategy that combines Amber Pruner with SmoothQuant (W8A8) \cite{xiao2023smoothquant}, termed Outstanding-sparse. This approach allows activation sparsity and weight quantization to stack effectively, significantly improving inference efficiency not only in the prefill phase but also showing promising acceleration in decoding. While current general-purpose hardware has limited support for such fine-grained sparsity, Outstanding-sparse highlights a forward-looking optimization path for LLM inference.

To the best of our knowledge, this is the first work exploring training-free N:M activation sparsity to accelerate LLM inference. Our main contributions are as follows:
\begin{itemize}
\item We propose a novel N:M activation sparsification algorithm that improves hardware efficiency without requiring retraining or fine-tuning, while maintaining stable performance across multiple tasks.
\item We introduce a heuristic layer skipping strategy tailored for activation sparsity, which effectively balances the pruning ratio and model quality.
\item We present a practical inference optimization algorithm that combines activation sparsification with weight quantization, demonstrating the feasibility of this direction and offering a concrete pathway for future software-hardware co-optimization.
\end{itemize}

\section{Related Work}

\subsection{Weight Sparsity}
N:M sparsity was first introduced by NVIDIA \cite{mishra2021accelerating} and applied in its Ampere architecture. One of the most typical applications of N:M sparsity is weight sparsity, especially in modules like linear projections, where sparse masks that conform to the N:M structure are predicted offline. A large body of research employs training-free heuristic algorithms, among which the most representative methods are SparseGPT \cite{frantar2023sparsegpt} and Wanda \cite{sun2023simple}. Subsequent studies have introduced various scoring strategies, such as distance \cite{zhang2023dynamic}, gradient \cite{das2023beyond}, entropy \cite{li2023sparse}, and low-rank approximations \cite{zhang2024oats}. Additionally, some research has explored learning-based approaches to obtain better sparse masks by combining lightweight learning mechanisms, such as MaskLLM \cite{fang2024maskllm} and ProxSparse \cite{liu2025proxsparse}.

While training-free methods offer advantages in deployment efficiency and scalability, they still face certain limitations in terms of accuracy. As a result, some studies have shifted toward training-aware sparsification methods, which explicitly guide the sparse structure during training to improve performance. These include methods such as SLoPe \cite{mozaffari2024slope}, Hu et al. \cite{hu2024accelerating}, S-STE \cite{hu2024s}, and AST \cite{huang2025pruning}. However, these methods typically come with higher training costs, limiting their widespread application in large-scale pre-trained models.

\subsection{Activation Sparsity}
The study of activation sparsity can be traced back to Li et al. \cite{li2022lazy}. They discovered that, in Transformer architectures, the output of the activation functions in the MLP modules often exhibits high sparsity, which further intensifies as the model size grows. Later, Dong et al. \cite{dong2023towards} systematically analyzed the inherent sparsity in Transformers, emphasizing that the choice of activation function plays a key role in the degree of activation sparsity.

To further explore activation sparsity, Liu et al. \cite{liu2023deja} proposed Déjà Vu, revealing significant contextual sparsity during the inference phase of LLMs. Mirzadeh et al. \cite{mirzadeh2023relu} advocated replacing complex activation functions such as GELU and SiLU with simpler ReLU-like functions to foster the emergence of sparsity. Similar explorations include ProSparse \cite{song2024prosparse} and Turbo Sparse \cite{song2024turbo}. Furthermore, Wang et al. \cite{wang2024q} and Haziza et al. \cite{haziza2025accelerating} showed that using Squared ReLU as an activation function naturally encourages stronger activation sparsity during training, laying the foundation for activation pruning. Zhang et al. \cite{zhang2025r} propose R-Sparse, which compresses linear layers by combining top-k activation sparsity with low-rank compensation for efficient inference.


\subsection{Combining N:M Sparsity with Quantization}
The integration of semi-structured sparsity with quantization strategies is a critical direction for improving the computational efficiency and memory utilization of LLMs. SDQ \cite{jeong2024sdq} was the first to merge N:M sparsity with W4A4/W8A8 quantization, effectively leveraging hardware's native support to achieve an optimal Pareto balance between model accuracy and computational acceleration. JSQ \cite{guo2024compressing} approached the problem from a joint optimization perspective, systematically exploring the co-design of 2:4/4:8 sparsity structures and quantization strategies. GQSA \cite{zeng2024gqsa} deeply couples 2:4 sparsity with per-group weight quantization, significantly improving computational and memory access efficiency while maintaining the accuracy of LLMs. Recently, Laborde et al. \cite{laborde2025semantic} investigated the synergistic effects between pruning and quantization, facilitating a more reasonable trade-off between model compression and semantic preservation. These studies collectively highlight the potential and practical value of combining structured sparsity and quantization techniques for model compression. 

\section{Methodology}
\begin{figure}[b]
\centering
\includegraphics[width=\columnwidth]{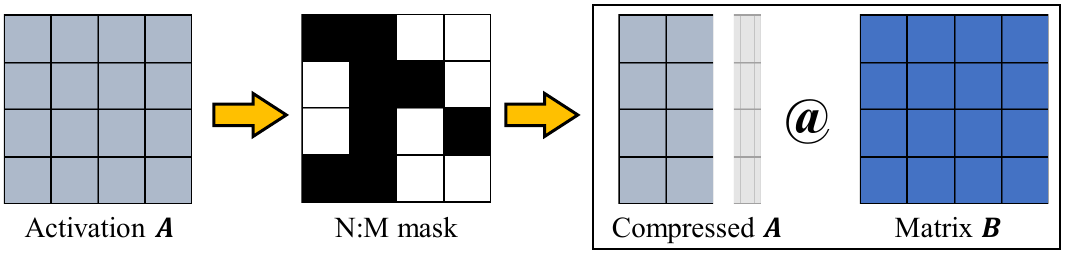}
\caption{Overview of Amber Pruner pipeline.}
\label{fig1}
\end{figure}

In this section, we introduce Amber Pruner, a training-free N:M activation sparsity algorithm tailored to accelerate linear projection computations in LLMs. Furthermore, we propose Outstanding-sparse, an approach that incorporates N:M activation sparsity into W8A8 quantized models for further efficiency gains. 
Figure \ref{fig1} illustrates the pipeline.

\subsection{Amber Pruner}
Amber Pruner refers to the structured N:M pruning of input activations in linear layers within LLMs, where the pruned activations are used as sparse inputs for subsequent computations. This form of structured sparsity can be interpreted as a semantics-preserving compression strategy. The resulting sparse activations are multiplied with dense weight matrices, forming a sparse-dense matrix multiplication (SpMM) scenario that enables hardware-level acceleration.

\subsubsection{Preliminary Observations}
Previous studies have shown that, in LLMs, activation values tend to exhibit more outliers than weights do, with extreme values (top \textless 1\%) often concentrated in specific channels \cite{xiao2023smoothquant, dettmers2208int8}. Building upon this, our empirical analysis reveals an additional observation: despite the fact that linear layers internally execute dense matrix multiplications, activation values contain significantly more elements close to zero compared to weight values.

\begin{figure}[t]
\centering
\includegraphics[width=\columnwidth]{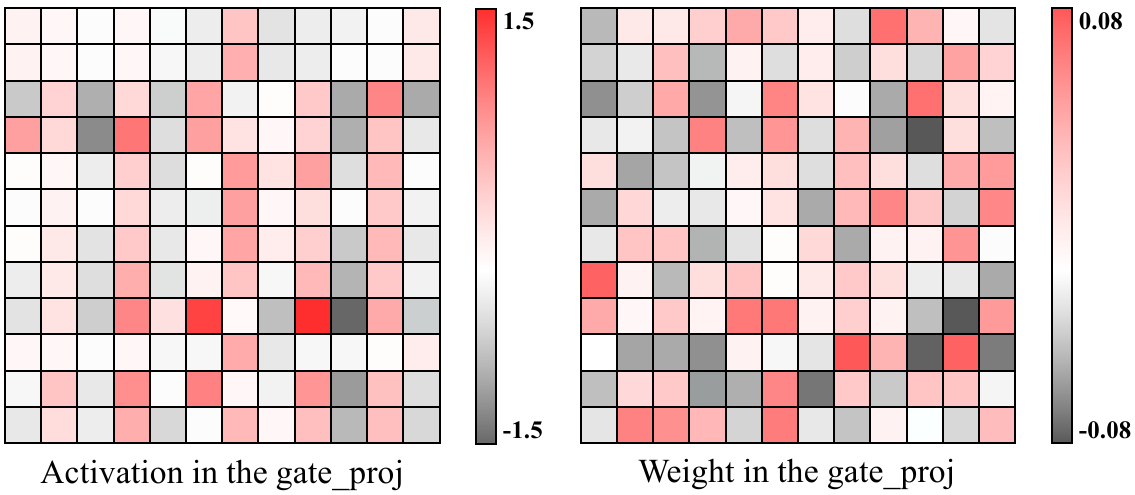}
\caption{Visual comparison of activations and weights in the gate projection.}
\label{fig2}
\end{figure}

As illustrated in Figure \ref{fig2}, roughly 50\% of activation values appear whiter (i.e., closer to zero within their min-max range), suggesting that activations are more amenable to structured sparsification. A comparison between activation and weight sparsity can be found in \textbf{Appendix A}. In contrast, weight tensors exhibit a more uniform distribution, lacking such sparsity characteristics. Motivated by these findings, we propose a straightforward yet effective strategy: applying top-k selection on activation elements to construct N:M structured sparse representations, which helps speed up inference and improve overall compute efficiency.

\subsubsection{Weight-Aware Scoring}
While directly applying top-k selection is simple and efficient, it can fail to preserve critical outputs, potentially degrading model accuracy. To address this, Wanda \cite{sun2023simple} proposes a straightforward pruning strategy that takes activations into account when pruning weights $W \in \mathbb{R}^{d_{out} \times d_{in}}$. The score is defined as
\begin{equation}
S_{ij} = \left | W_{ij} \right | \cdot f\left ( X_{:,j} \right ) =\left | W_{ij} \right | \cdot {\left \| X_{:,j} \right \|}_2. \label{eq1}
\end{equation}
In contrast, since our focus is on pruning the activations $X$, we reverse this formulation by considering statistics of the weight matrix. A Wanda-like approach (see \textbf{Appendix B}) gives the scoring function as
\begin{equation}
S_{ij} = \left | X_{ij} \right | \cdot f\left ( W_{:,j} \right ) = \left | X_{ij} \right | \cdot \left ( \frac{{\left \| W_{:,j} \right \|}_2 }{\min_{k} {\left \| W_{:,k} \right \|}_2 }  \right ).  \label{eq2} 
\end{equation}

Note that in many models, certain weight channels have a limited dynamic range. Using raw L2 norms as weighting factors can result in excessively small scores, risking numerical underflow in low-precision inference and reducing pruning effectiveness. To address this, we use a simple remedy: minimum normalization of channel norms to ensure stability and retain key activations, as shown in Equation \ref{eq2}.

Although L2-norm-based scoring reflects the relative importance of different channels, weight tensors often exhibit concentrated and low-variance distributions. As a result, relying solely on raw norms may blur distinctions among critical boundary channels. To overcome this limitation, we propose a Robust-Norm Scoring mechanism that improves both the stability and discriminative power of the scoring function by applying normalization to the weight values. The method consists of the following key steps:
\begin{enumerate}
\item \textbf{Outlier Removal}. To reduce the influence of extreme values, weight $W$ outside the 0.5th–99.5th percentiles are discarded:
\begin{equation}
W=\left \{ \omega_k ~|~ Q_{0.005}\left ( W \right )  \le \omega_k \le Q_{0.995} \left ( W \right ) \right \} ; \label{eq3} 
\end{equation}
\item \textbf{Normalization}. The remaining weights are standardized using their mean and variance:
\begin{equation}
\hat{W}_{ij} = \frac{W_{ij} - \mathbb{E}\left [ W \right ] }{\sqrt{{\rm Var} \left [ W \right ] } }; \label{eq4} 
\end{equation}
\item \textbf{Channel-wise Scoring}. For each channel $\hat{W}_{:,j}$, the L2 norm is computed, and final scores are assigned to activation elements via a Wanda-like rule (see Equation 2):
\begin{equation}
S_{ij}^* = \left | X_{ij} \right | \cdot f\left ( \hat{W}_{:,j} \right ) . \label{eq5} 
\end{equation}
\end{enumerate}

Robust-Norm Scoring enhances local scoring resolution via numerical normalization. This fine-grained adjustment helps identify boundary-critical channels without violating OBS optimality constraints \cite{hassibi1993optimal}, boosting end-to-end accuracy without noise.

Since model weights remain fixed during inference, the Robust-Norm Scoring coefficients can be precomputed offline prior to deployment. These scaling factors are stored as auxiliary weights and integrated into the model without the need for additional training or fine-tuning. The associated overhead is minimal: it constitutes less than 0.05\% of the total model size and has a negligible impact on storage and memory usage. Moreover, channel-wise scaling can be incorporated into the computation graph via operator fusion techniques, enabling efficient and transparent acceleration during inference.

\subsubsection{Layer Skipping Strategy}
Given that activation distributions vary significantly across linear projections (see \textbf{Appendix C}), some layers are more amenable to sparsification than others. To mitigate potential performance degradation caused by indiscriminate activation pruning, we introduce a layer skipping strategy that selectively bypasses sparsification for sensitive layers.

Skipping sensitivity-prone layers is a well-established practice in model pruning. Gromov et al. \cite{gromov2024unreasonable} assessed layer-wise semantic similarity and adopted a backward removal strategy to meet pruning targets. Men et al. \cite{men2024shortgpt} used cosine similarity between adjacent layers to estimate pruning risk. Zhang et al. \cite{zhang2024finercut} applied Jensen-Shannon divergence to capture changes in output logits, highlighting the redundancy of deeper self-attention layers. Ma et al. \cite{ma2024dynamic} introduced an L2 norm-based sensitivity metric to quantify the impact of activation pruning. Liang et al. \cite{liang2025seap} proposed SEAP, a task-aware method that activates only the most relevant neurons, emphasizing the importance of layers near the output for language modeling.

Inspired by sensitivity analysis methods for weight sparsity, we propose a method tailored for activation sparsity that measures the functional deviation caused by pruning. For example, consider a query projection layer where the input activation is $X$, and the original forward path is
\begin{align}
Q &= X W_q, K = X W_k, V = X W_v,  \notag \\
\tilde{Q} &= {\rm RoPE}(Q), \tilde{K} = {\rm RoPE}(K), \notag \\
Y &= {\rm MLP}\left ( {\rm Attn} \left ( \tilde{Q}, \tilde{K}, V \right )  \right ). \label{eq6} 
\end{align}
If structured N:M pruning is applied to obtain a sparse input $X'$, the forward path becomes
\begin{align}
Q' &= X' W_q, \tilde{Q'} = {\rm RoPE}(Q') \notag \\
Y' &= {\rm MLP}\left ( {\rm Attn} \left ( \tilde{Q'}, \tilde{K}, V \right )  \right ). \label{eq7}
\end{align}
To quantify the deviation introduced by sparse activations, we define a relative perturbation error
\begin{equation}
e_q\left ( Y, Y' \right ) =\frac{{\left \| Y-Y' \right \|}_2 }{{\left \| Y \right \| }_2 + \varepsilon },  \label{eq8}
\end{equation}
where $\varepsilon$ is a small constant introduced to ensure numerical stability. The metric $e_q$ quantifies how activation sparsity in the query projection affects the downstream output in a full forward pass. The sensitivity of different linear projections can be referred to in \textbf{Appendix D}.

In short, the proposed sensitivity analysis provides an efficient way to estimate how pruning affects the representational capacity of linear projections. However, empirical end-to-end evaluation is still essential to ensure accuracy, especially for layers that are sensitive to small perturbations.

\subsection{Outstanding-sparse}
To evaluate the generalizability of Amber Pruner for low-bit deployment, we propose Outstanding-sparse, which integrates activation sparsification with quantization. Specifically, we adopt the widely used W8A8 quantization method based on SmoothQuant. As proposed in \cite{xiao2023smoothquant}, SmoothQuant introduces a channel-wise scaling strategy to shift outliers between activations and weights, with this process controlled by a hyperparameter $\alpha \in \left [ 0, 1 \right ] $. The scaling factor is defined as
\begin{equation}
s_j = \frac{{{\rm max}\left ( \left | X_{:,j} \right |  \right ) }^{\alpha}}{{{\rm max}\left ( \left | W_{:,j} \right |  \right ) }^{1 - \alpha}}, j=1,2,...,{\rm dim}. \label{eq9}
\end{equation}
Here, a larger $\alpha$ pushes outliers into the weights, compressing the activation range and making activations easier to quantize; a smaller $\alpha$ does the reverse.

\begin{figure}[t]
\centering
\includegraphics[width=\columnwidth]{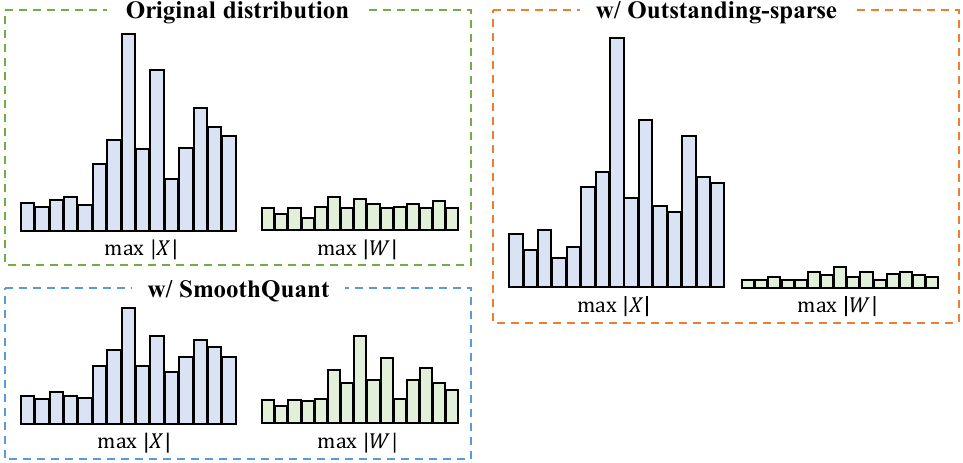}
\caption{Visual comparison of vanilla SmoothQuant and Outstanding-sparse (with $\alpha = 0.10$).}
\label{fig3}
\end{figure}

Our empirical observations show that Amber Pruner performs better in sparsification when the activation range is expanded, as this helps reveal structured sparsity patterns more effectively. To adapt SmoothQuant to this sparsity-aware setting, we redefine the scaling factor in Outstanding-sparse as $\hat{s}_j = {1}/{s_j}$. Unlike the original formulation, $\hat{s}_j$ explicitly expands the activation range under the same $\alpha$. As illustrated in Figure \ref{fig3}, we favor a small $\alpha$ to enhance the synergy between quantization and sparsification.

\section{Experiments}
In this section, we evaluate Amber Pruner and Outstanding-sparse across models and tasks, and summarize key results.
\subsection{Amber Pruner}
\subsubsection{Experimental Setup}

\begin{table*}[htbp]
\centering
\fontsize{8pt}{8pt}\selectfont
\caption{Evaluation results of Amber Pruner on Zero-shot tasks.}
\begin{tabular}{lcl|ccccccccc|cc} 
\toprule
\textbf{Model}                 & \textbf{Rt.}          & \textbf{Settings}   & \textbf{AC} & \textbf{AE} & \textbf{BQ} & \textbf{MMLU} & \textbf{CEVAL} & \textbf{OBQA} & \textbf{PIQA} & \textbf{RTE} & \textbf{WG} & \textbf{Avg.} & \textbf{Drop}  \\ 
\midrule
\multirow{12}{*}{\rotatebox{80}{LLaMA3.1-8B}}  & -     & Bfloat16            & 0.5196      & 0.8178      & 0.8416      & 0.6807        & -              & 0.3340        & 0.8003        & 0.6823       & 0.7411      & 0.6772        &  -             \\
\cmidrule{2-14}
                               & \multirow{3}{*}{2:4}  & Naïve top-k         & 0.3874      & 0.7184      & 0.7590      & 0.4919        & -              & 0.2640        & 0.7285        & 0.6029       & 0.6417      & 0.5742        & -10.3\%        \\
                               &                       & Amber-P (l.s.)      & 0.4829      & 0.8009      & 0.8355      & 0.6047        & -              & 0.3260        & 0.7786        & 0.6426       & 0.7324      & 0.6505        & -2.7\%         \\
                               &                       & Amber-P (all)       & 0.4940      & 0.7980      & 0.8352      & 0.6009        & -              & 0.3260        & 0.7824        & 0.6715       & 0.7230      & 0.6539        & -2.3\%         \\ 
\cmidrule{2-14}
                               & \multirow{3}{*}{4:8}  & Naïve top-k         & 0.4249      & 0.7597      & 0.7939      & 0.5502        & -              & 0.2720        & 0.7497        & 0.6101       & 0.6677      & 0.6035        & -7.4\%         \\
                               &                       & Amber-P (l.s.)      & 0.4915      & 0.7955      & 0.8330      & 0.6286        & -              & 0.3300        & 0.7835        & 0.6715       & 0.7411      & 0.6593        & -1.8\%         \\
                               &                       & Amber-P (all)       & 0.4915      & 0.8030      & 0.8358      & 0.6300        & -              & 0.3300        & 0.7818        & 0.6715       & 0.7277      & 0.6589        & -1.8\%         \\ 
\cmidrule{2-14}
                               & \multirow{3}{*}{8:16} & Naïve top-k         & 0.4266      & 0.7694      & 0.8052      & 0.5742        & -              & 0.3240        & 0.7682        & 0.6282       & 0.6890      & 0.6231        & -5.4\%         \\
                               &                       & Amber-P (l.s.)      & 0.5043      & 0.8043      & 0.8401      & 0.6386        & -              & 0.3340        & 0.7905        & 0.6751       & 0.7230      & 0.6637        & -1.4\%         \\
                               &                       & Amber-P (all)       & 0.5145      & 0.8030      & 0.8443      & 0.6369        & -              & 0.3320        & 0.7856        & 0.7004       & 0.7459      & 0.6703        & -0.7\%         \\ 
\midrule
\multirow{12}{*}{\rotatebox{80}{Qwen2-7B}}     & -     & Bfloat16            & 0.5085      & 0.8043      & 0.8541      & 0.8188        & 0.8135         & 0.3460        & 0.7954        & 0.7870       & 0.6969      & 0.7138        & -              \\
\cmidrule{2-14}
                               & \multirow{3}{*}{2:4}  & Naïve top-k         & 0.4360      & 0.7567      & 0.8284      & 0.6080        & 0.6308         & 0.2960        & 0.7573        & 0.7220       & 0.6448      & 0.6311        & -8.3\%         \\
                               &                       & Amber-P (l.s.)      & 0.5213      & 0.8047      & 0.8563      & 0.7399        & 0.7511         & 0.3480        & 0.7862        & 0.7690       & 0.6993      & 0.6973        & -1.7\%         \\
                               &                       & Amber-P (all)       & 0.5230      & 0.7997      & 0.8517      & 0.7405        & 0.7437         & 0.3400        & 0.7878        & 0.7473       & 0.7096      & 0.6937        & -2.0\%         \\ 
\cmidrule{2-14}
                               & \multirow{3}{*}{4:8}  & Naïve top-k         & 0.4650      & 0.7753      & 0.8370      & 0.6716        & 0.6961         & 0.3000        & 0.7682        & 0.7798       & 0.6740      & 0.6630        & -5.1\%         \\
                               &                       & Amber-P (l.s.)      & 0.5077      & 0.7997      & 0.8554      & 0.7621        & 0.7675         & 0.3480        & 0.7840        & 0.7690       & 0.7245      & 0.7020        & -1.2\%         \\
                               &                       & Amber-P (all)       & 0.5111      & 0.8013      & 0.8566      & 0.7619        & 0.7667         & 0.3540        & 0.7873        & 0.7617       & 0.7032      & 0.7004        & -1.3\%         \\ 
\cmidrule{2-14}
                               & \multirow{3}{*}{8:16} & Naïve top-k         & 0.4787      & 0.7841      & 0.8462      & 0.7067        & 0.7028         & 0.3100        & 0.7688        & 0.7473       & 0.6748      & 0.6688        & -4.5\%         \\
                               &                       & Amber-P (l.s.)      & 0.5043      & 0.7984      & 0.8572      & 0.7740        & 0.7637         & 0.3440        & 0.7960        & 0.7581       & 0.6993      & 0.6994        & -1.4\%         \\
                               &                       & Amber-P (all)       & 0.5068      & 0.8077      & 0.8609      & 0.7744        & 0.7682         & 0.3500        & 0.7884        & 0.7834       & 0.7127      & 0.7058        & -0.8\%         \\ 
\midrule
\multirow{9}{*}{\rotatebox{80}{Qwen3-30B-A3B}} & -     & Bfloat16            & 0.5282      & 0.7959      & 0.8865      & 0.8065        & 0.8410         & 0.3420        & 0.7933        & 0.8195       & 0.7135      & 0.7252        & -              \\
\cmidrule{2-14}
                               & \multirow{2}{*}{2:4}  & Naïve top-k         & 0.4488      & 0.7588      & 0.8554      & 0.6907        & 0.7400         & 0.3100        & 0.7476        & 0.7726       & 0.6551      & 0.6643        & -6.1\%         \\
                               &                       & Amber-P (l.s.)      & 0.5077      & 0.7761      & 0.8774      & 0.7730        & 0.7957         & 0.3260        & 0.7797        & 0.7942       & 0.7064      & 0.7040        & -2.1\%         \\ 
\cmidrule{2-14}
                               & \multirow{2}{*}{4:8}  & Naïve top-k         & 0.4676      & 0.7572      & 0.8682      & 0.7239        & 0.7563         & 0.3160        & 0.7671        & 0.8014       & 0.6661      & 0.6804        & -4.5\%         \\
                               &                       & Amber-P (l.s.)      & 0.5119      & 0.7778      & 0.8780      & 0.7791        & 0.8105         & 0.3440        & 0.7862        & 0.8195       & 0.7222      & 0.7144        & -1.1\%         \\ 
\cmidrule{2-14} 
                               & \multirow{2}{*}{8:16} & Naïve top-k         & 0.4863      & 0.7744      & 0.8667      & 0.7374        & 0.7571         & 0.3420        & 0.7677        & 0.8303       & 0.6835      & 0.6939        & -3.1\%         \\
                               &                       & Amber-P (l.s.)      & 0.5401      & 0.7950      & 0.8872      & 0.7854        & 0.8232         & 0.3260        & 0.7938        & 0.8267       & 0.7174      & 0.7216        & -0.4\%         \\
\bottomrule
\end{tabular}
\label{tab1}
\end{table*}

We assess Amber Pruner on two representative categories of LLMs:
\begin{itemize}
\item \textbf{Dense models}: LLaMA3.1-8B-Instruct \cite{grattafiori2024llama} and Qwen2-7B-Instruct \cite{team2024qwen2};
\item \textbf{Sparse Mixture-of-Experts (MoE) models}: represented by Qwen3-30B-A3B \cite{yang2025qwen3}.
\end{itemize}
We evaluate three commonly used structured sparsity ratios: 2:4, 4:8, and 8:16. Model performance is measured on a range of downstream tasks, including:
\begin{itemize}
\item \textbf{Zero-shot tasks}: including ARC-Challenge (AC) \cite{clark2018think}, ARC-Easy (AE), BoolQ (BQ) \cite{clark2019boolq}, OpenBookQA (OBQA) \cite{OpenBookQA2018}, PIQA \cite{Bisk2020}, RTE \cite{wang2018glue}, Winogrande (WG), and MMLU \cite{hendryckstest2021} for the English-based LLaMA model, as well as CMMLU \cite{li2023cmmlu} and CEVAL \cite{huang2023ceval} for the Chinese-based Qwen model.
\item \textbf{Few-shot tasks}: represented by GSM8K (5-shot) \cite{cobbe2021gsm8k};
\item \textbf{Long-context comprehension tasks}: selected from the LongBench benchmark \cite{bai2023longbench}, e.g., TriviaQA.
\end{itemize}


Pruning is applied to the linear projection submodules in the prefill stage, including q\_proj, k\_proj, v\_proj, and o\_proj in the attention module, as well as gate\_proj, up\_proj, and down\_proj in the MLP/MoE module.

To balance computational efficiency and accuracy, we apply a skipping strategy guided by sensitivity analysis. Specifically, due to the use of Grouped Query Attention (GQA), the computational burden of the k\_proj and v\_proj modules is relatively low, and they are directly marked as non-prunable. Beyond that, sensitivity scores averaged across all layers indicate that o\_proj and up\_proj are critical and should also be preserved. In contrast, down\_proj consistently exhibits the lowest sensitivity and is thus pruned in all layers. Meanwhile, q\_proj and gate\_proj display varying sensitivity across layers, making them suitable candidates for selective pruning based on performance requirements:
\begin{itemize}
\item \textbf{LLaMA3.1-8B}: q\_proj and gate\_proj are skipped in layers 19, 21, 28, 30, and 31, leading to 56.1\% of total linear computation being accelerated.
\item \textbf{Qwen2-7B}: q\_proj and gate\_proj are skipped in layers 0, 6, 23, 26, and 27, resulting in 57.6\% of total linear computation being accelerated.
\item \textbf{Qwen3-30B-A3B}: q\_proj and gate\_proj are skipped in layers 41, 46, and 47, yielding 56.9\% acceleration coverage in terms of overall linear computation.
\end{itemize}

\begin{table*}[ht]
\centering
\fontsize{8pt}{8pt}\selectfont
\caption{Evaluation results of Outstanding-sparse on Zero-shot tasks.}
\begin{tabular}{lcl|ccccccccc|cc} 
\toprule
\textbf{Model}                 & \textbf{Rt.}          & \textbf{Settings}   & \textbf{AC} & \textbf{AE} & \textbf{BQ} & \textbf{MMLU} & \textbf{CEVAL} & \textbf{OBQA} & \textbf{PIQA} & \textbf{RTE} & \textbf{WG} & \textbf{Avg.} & \textbf{Drop}  \\ 
\midrule
\multirow{12}{*}{\rotatebox{80}{LLaMA3.1-8B}}  & -     & SQ-W8A8             & 0.5128      & 0.8148      & 0.8395      & 0.6760        & -              & 0.3320        & 0.7987        & 0.6968       & 0.7348      & 0.6757        & -              \\
\cmidrule{2-14}
                               & \multirow{3}{*}{2:4}  & Naïve top-k         & 0.3686      & 0.7037      & 0.7575      & 0.4743        & -              & 0.2320        & 0.7236        & 0.5993       & 0.6456      & 0.5631        & -11.3\%        \\
                               &                       & O-sparse (l.s.)     & 0.4932      & 0.7904      & 0.8257      & 0.6186        & -              & 0.3040        & 0.7688        & 0.6751       & 0.6961      & 0.6465        & -2.9\%         \\
                               &                       & O-sparse (all)      & 0.4821      & 0.7967      & 0.8287      & 0.6186        & -              & 0.3040        & 0.7748        & 0.6968       & 0.6953      & 0.6496        & -2.6\%         \\ 
\cmidrule{2-14}
                               & \multirow{3}{*}{4:8}  & Naïve top-k         & 0.4078      & 0.7471      & 0.7896      & 0.5349        & -              & 0.2700        & 0.7563        & 0.6137       & 0.6693      & 0.5986        & -7.7\%         \\
                               &                       & O-sparse (l.s.)     & 0.4923      & 0.7938      & 0.8306      & 0.6354        & -              & 0.3240        & 0.7813        & 0.7040       & 0.7080      & 0.6587        & -1.7\%         \\
                               &                       & O-sparse (all)      & 0.4974      & 0.8060      & 0.8312      & 0.6350        & -              & 0.3320        & 0.7786        & 0.7004       & 0.7324      & 0.6641        & -1.2\%         \\ 
\cmidrule{2-14}
                               & \multirow{3}{*}{8:16} & Naïve top-k         & 0.4352      & 0.7643      & 0.8067      & 0.5672        & -              & 0.2920        & 0.7633        & 0.6498       & 0.6827      & 0.6202        & -5.7\%         \\
                               &                       & O-sparse (l.s.)     & 0.4991      & 0.8068      & 0.8297      & 0.6388        & -              & 0.3280        & 0.7786        & 0.6787       & 0.7127      & 0.6590        & -1.7\%         \\
                               &                       & O-sparse (all)      & 0.5137      & 0.8064      & 0.8349      & 0.6389        & -              & 0.3240        & 0.7824        & 0.7040       & 0.7167      & 0.6654        & -1.1\%         \\ 
\midrule
\multirow{12}{*}{\rotatebox{80}{Qwen2-7B}}     & -     & SQ-W8A8             & 0.5060      & 0.8051      & 0.8538      & 0.8160        & 0.8076         & 0.3440        & 0.7922        & 0.7870       & 0.6882      & 0.7111        & -              \\
\cmidrule{2-14}
                               & \multirow{3}{*}{2:4}  & Naïve top-k         & 0.4497      & 0.7412      & 0.8183      & 0.6019        & 0.6300         & 0.3000        & 0.7486        & 0.7762       & 0.6646      & 0.6367        & -7.4\%         \\
                               &                       & O-sparse (l.s.)     & 0.5102      & 0.7900      & 0.8483      & 0.7129        & 0.7184         & 0.3340        & 0.7878        & 0.7509       & 0.6827      & 0.6817        & -2.9\%         \\
                               &                       & O-sparse (all)      & 0.5119      & 0.7917      & 0.8544      & 0.7095        & 0.7214         & 0.3440        & 0.7813        & 0.7473       & 0.6882      & 0.6833        & -2.8\%         \\ 
\cmidrule{2-14}
                               & \multirow{3}{*}{4:8}  & Naïve top-k         & 0.4667      & 0.7875      & 0.8346      & 0.6659        & 0.6568         & 0.2980        & 0.7644        & 0.7581       & 0.6440      & 0.6529        & -5.8\%         \\
                               &                       & O-sparse (l.s.)     & 0.5068      & 0.8018      & 0.8572      & 0.7619        & 0.7571         & 0.3400        & 0.7900        & 0.7798       & 0.6953      & 0.6989        & -1.2\%         \\
                               &                       & O-sparse (all)      & 0.5068      & 0.8039      & 0.8550      & 0.7597        & 0.8143         & 0.3380        & 0.7949        & 0.7762       & 0.7040      & 0.7059        & -0.5\%         \\ 
\cmidrule{2-14}
                               & \multirow{3}{*}{8:16} & Naïve top-k         & 0.4761      & 0.7896      & 0.8401      & 0.7001        & 0.6954         & 0.3380        & 0.7682        & 0.7834       & 0.6867      & 0.6753        & -3.6\%         \\
                               &                       & O-sparse (l.s.)     & 0.4991      & 0.7921      & 0.8575      & 0.7737        & 0.7660         & 0.3420        & 0.7884        & 0.7726       & 0.6756      & 0.6963        & -1.5\%         \\
                               &                       & O-sparse (all)      & 0.5068      & 0.8039      & 0.8538      & 0.7700        & 0.8143         & 0.3380        & 0.7949        & 0.7762       & 0.7040      & 0.7069        & -0.4\%         \\ 
\midrule
\multirow{9}{*}{\rotatebox{80}{Qwen3-30B-A3B}} & -     & SQ-W8A8             & 0.5282      & 0.7942      & 0.8881      & 0.8100        & 0.8373         & 0.3380        & 0.7954        & 0.8267       & 0.6946      & 0.7236        & -              \\
\cmidrule{2-14}
                               & \multirow{2}{*}{2:4}  & Naïve top-k         & 0.4718      & 0.7382      & 0.8431      & 0.6993        & 0.7192         & 0.3000        & 0.7628        & 0.7690       & 0.6409      & 0.6605        & -6.3\%         \\
                               &                       & O-sparse (l.s.)     & 0.5145      & 0.7736      & 0.8801      & 0.7765        & 0.8001         & 0.3420        & 0.7807        & 0.8195       & 0.8001      & 0.7087        & -1.5\%         \\ 
\cmidrule{2-14}
                               & \multirow{2}{*}{4:8}  & Naïve top-k         & 0.4872      & 0.7593      & 0.8651      & 0.7268        & 0.7585         & 0.3220        & 0.7639        & 0.7906       & 0.6685      & 0.6824        & -4.1\%         \\
                               &                       & O-sparse (l.s.)     & 0.5034      & 0.7887      & 0.8804      & 0.7849        & 0.7873         & 0.3320        & 0.7957        & 0.8195       & 0.6953      & 0.7097        & -1.4\%         \\ 
\cmidrule{2-14}
                               & \multirow{2}{*}{8:16} & Naïve top-k         & 0.4795      & 0.7677      & 0.8648      & 0.7449        & 0.7764         & 0.3340        & 0.7650        & 0.7906       & 0.6835      & 0.6896        & -3.4\%         \\
                               &                       & O-sparse (l.s.)     & 0.5256      & 0.7879      & 0.8807      & 0.7833        & 0.8016         & 0.3520        & 0.7824        & 0.8123       & 0.7119      & 0.7153        & -0.8\%         \\
\bottomrule
\end{tabular}
\label{tab2}
\end{table*}

\begin{figure*}[!ht]
\centering
\begin{subfigure}{0.22\textwidth}
    \includegraphics[width=\linewidth]{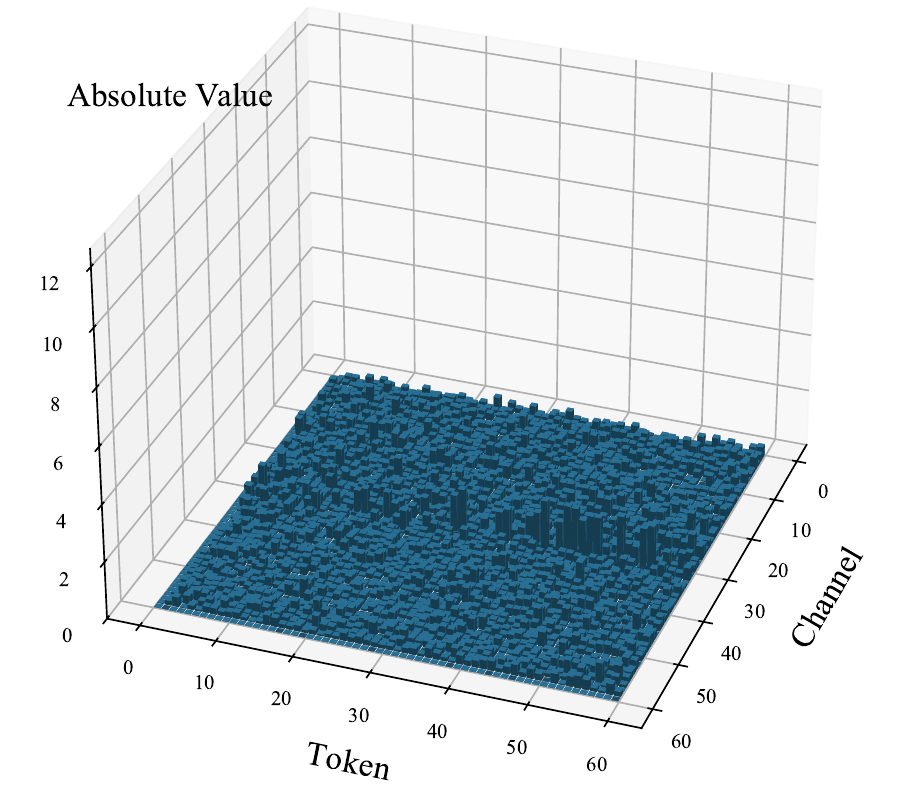}
    \caption{Activation (Pre)}
    \label{fig5:fig5a}
\end{subfigure}
\hfill
\begin{subfigure}{0.22\textwidth}
    \includegraphics[width=\linewidth]{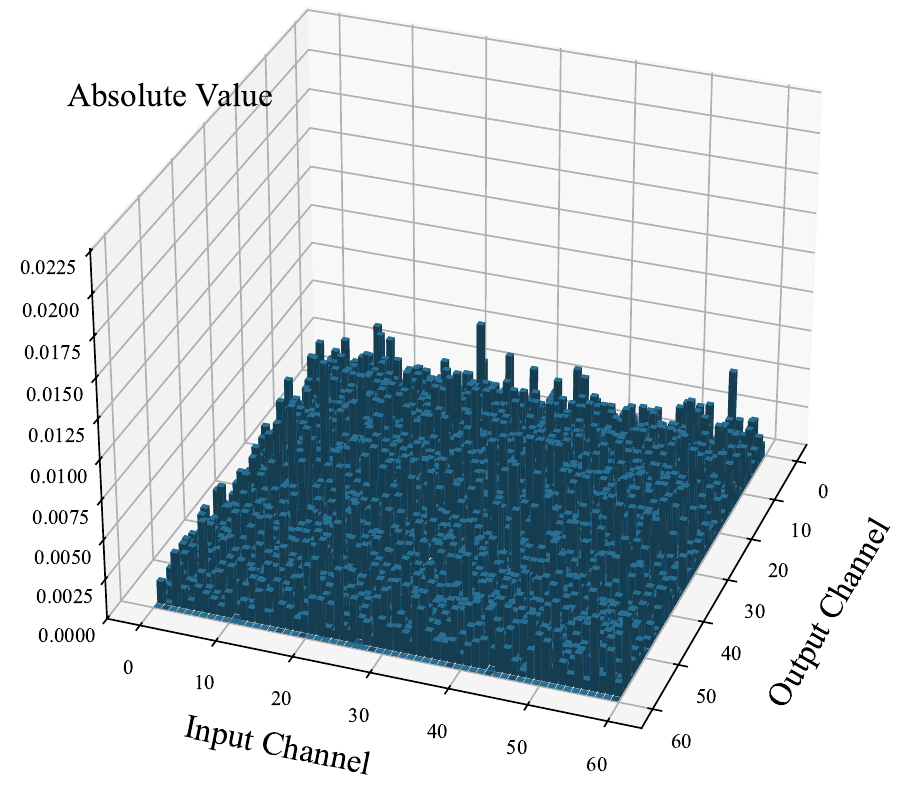}
    \caption{Weight (Pre)}
    \label{fig5:fig5b}
\end{subfigure}
\hfill
\begin{subfigure}{0.22\textwidth}
    \includegraphics[width=\linewidth]{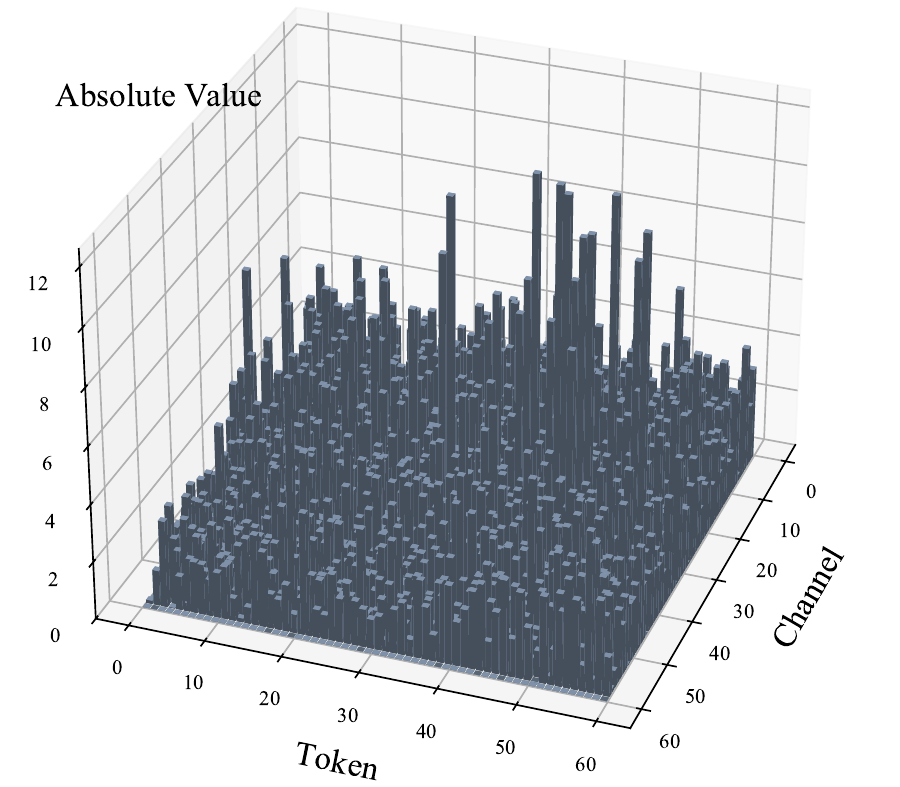}
    \caption{Activation (Post)}
    \label{fig5:fig5c}
\end{subfigure}
\hfill
\begin{subfigure}{0.22\textwidth}
    \includegraphics[width=\linewidth]{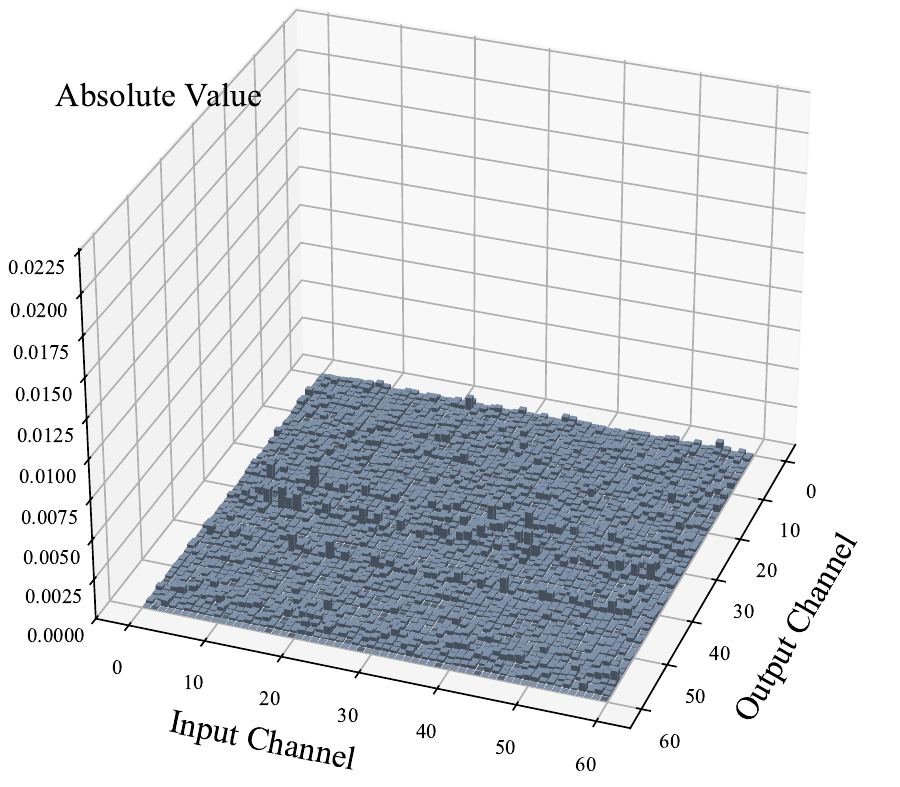}
    \caption{Weight (Post)}
    \label{fig5:fig5d}
\end{subfigure}
\caption{Impact of Outstanding-sparse on activation and weight distributions (Pre/Post adjustment).}
\label{fig5}
\end{figure*}

\subsubsection{Results}
Table \ref{tab1} presents the results of Amber Pruner on Zero-shot tasks. We compare these baselines and variants:
\begin{itemize}
\item \textbf{Bfloat16}: Full-precision inference used as the baseline.
\item \textbf{Naïve top-k}: Prunes activations based solely on magnitude, representing a general activation sparsity approach \cite{wang2024q, liu2024training}.
\item \textbf{Amber-P (l.s.)}: Amber Pruner with only Layer Skipping (l.s.) enabled.
\item \textbf{Amber-P (all)}: Full version combining Robust-Norm Scoring (r.s.) and Layer Skipping (l.s.).
\end{itemize}

All experiments were conducted on a platform with 8$\times$ Ascend 910B processors. Note that Robust-Norm Scoring is not applicable to MoE models, since tokens are dynamically routed to different experts. Table \ref{tab1} presents several important findings:
\begin{itemize}
\item \textbf{Effect of M in N:M Sparsity}: As the M value increases in N:M sparsity, activation pruning retains more semantic information, leading to improved performance on downstream tasks. The 8:16 configuration is particularly effective, typically resulting in less than 1\% accuracy loss in Zero-shot evaluations.
\item \textbf{Benefits of Robust-Norm Scoring}: Our Robust-Norm Scoring further improves accuracy, especially under 8:16 sparsity, reducing Zero-shot accuracy loss to below 1\%.
\item \textbf{Strong Generalization to MoE Models}: Despite the relatively small volume of activated neurons in MoE models (e.g., only 3B in our experiments), Amber Pruner still delivers competitive performance, demonstrating robust generalizability across model architectures.
\end{itemize}

Furthermore, Layer Skipping (l.s.) proves to be a valuable technique. For applications requiring higher accuracy, selecting additional layers or modules to skip is often beneficial. As shown in Table \ref{tab1}, across a broad range of models, Amber Pruner effectively accelerates over 55\% of linear projection computations by pruning less sensitive components, while maintaining strong performance.

\subsection{Outstanding-sparse}
\subsubsection{Experimental Setup}

\begin{table*}[t]
\centering
\fontsize{8pt}{8pt}\selectfont 
\caption{Few-shot and LongBench results. Baselines: Bfloat16 (Amber Pruner), SQ-W8A8 (Outstanding-sparse).}
\begin{tabular}{lcl|cc|cc|cc|cc}
\toprule
                                  &                       &                            & \multicolumn{4}{c|}{\textbf{Amber Pruner}}                                       & \multicolumn{4}{c}{\textbf{Outstanding-sparse}} \\
                                  &                       &                            & \multicolumn{2}{c|}{\textbf{Few-shot}} & \multicolumn{2}{c|}{\textbf{Longbench}} & \multicolumn{2}{c|}{\textbf{Few-shot}} & \multicolumn{2}{c}{\textbf{Longbench}} \\
\cmidrule{4-11}
\textbf{Model}                    & \textbf{Rt.}          & \textbf{Settings}          & GSM8K            & Drop            & Avg.      & Drop               & GSM8K            & Drop            & Avg.       & Drop   \\
\midrule
\multirow{12}{*}{\rotatebox{80}{LLaMA3.1-8B}}        & -  & Baseline                   & 0.8036           & -               & 0.3056    & -                  & 0.7748           & -               & 0.3001     & -      \\
\cmidrule{2-11}
                                  & \multirow{3}{*}{2:4}  & Naïve top-k                & 0.6513           & -15.3\%         & 0.1937    & -11.2\%            & 0.5678           & -20.7\%         & 0.2116     & -8.9\% \\
                                  &                       & Proposed (l.s.)            & 0.7832           & -2.4\%          & 0.2933    & -1.2\%             & 0.7460           & -2.9\%          & 0.2799     & -2.0\% \\
                                  &                       & Proposed (all)             & 0.7976           & -0.6\%          & 0.2889    & -1.7\%             & 0.7672           & -0.8\%          & 0.2956     & -0.6\% \\
\cmidrule{2-11}
                                  & \multirow{3}{*}{4:8}  & Naïve top-k                & 0.7392           & -6.4\%          & 0.2527    & -5.3\%             & 0.6535           & -12.1\%         & 0.2723     & -2.8\% \\
                                  &                       & Proposed (l.s.)            & 0.7854           & -0.2\%          & 0.3022    & -0.3\%             & 0.7582           & -1.7\%          & 0.2974     & -0.3\% \\
                                  &                       & Proposed (all)             & 0.8120           & +0.8\%          & 0.2992    & -0.6\%             & 0.7779           & +0.3\%          & 0.2916     & -0.9\% \\
\cmidrule{2-11}
                                  & \multirow{3}{*}{8:16} & Naïve top-k                & 0.7635           & -4.0\%          & 0.2672    & -3.8\%             & 0.7339           & -4.1\%          & 0.2906     & -1.0\% \\
                                  &                       & Proposed (l.s.)            & 0.8014           & -0.2\%          & 0.3143    & +0.9\%             & 0.7657           & -0.9\%          & 0.2953     & -0.5\% \\
                                  &                       & Proposed (all)             & 0.8127           & +0.9\%          & 0.3091    & +0.3\%             & 0.7741           & -0.1\%          & 0.3017     & +0.2\% \\
\midrule
\multirow{12}{*}{\rotatebox{80}{Qwen2-7B}}           & -  & Baseline                   & 0.7907           & -               & 0.3133    & -                  & 0.7733           & -               & 0.3001     & -      \\
\cmidrule{2-11}
                                  & \multirow{3}{*}{2:4}  & Naïve top-k                & 0.7642           & -2.7\%          & 0.2366    & -7.7\%             & 0.7589           & -1.4\%          & 0.2274     & -7.3\% \\
                                  &                       & Proposed (l.s.)            & 0.7846           & -0.6\%          & 0.3111    & -0.2\%             & 0.7801           & +0.7\%          & 0.3020     & +0.2\% \\
                                  &                       & Proposed (all)             & 0.7771           & -1.4\%          & 0.3046    & -0.9\%             & 0.7784           & +0.5\%          & 0.3042     & +0.4\% \\
\cmidrule{2-11}
                                  & \multirow{3}{*}{4:8}  & Naïve top-k                & 0.7702           & -2.1\%          & 0.2507    & -6.3\%             & 0.7665           & -0.7\%          & 0.2607     & -3.9\% \\
                                  &                       & Proposed (l.s.)            & 0.7794           & -1.1\%          & 0.3110    & -0.2\%             & 0.7786           & +0.5\%          & 0.3035     & +0.4\% \\
                                  &                       & Proposed (all)             & 0.7741           & -1.7\%          & 0.3096    & -0.4\%             & 0.7793           & +0.3\%          & 0.3052     & +0.5\% \\
\cmidrule{2-11}
                                  & \multirow{3}{*}{8:16} & Naïve top-k                & 0.7604           & -3.0\%          & 0.2718    & -4.2\%             & 0.7627           & -1.1\%          & 0.2614     & -3.9\% \\
                                  &                       & Proposed (l.s.)            & 0.7854           & -0.5\%          & 0.3121    & -0.1\%             & 0.7801           & +0.7\%          & 0.3004     & +0.0\% \\
                                  &                       & Proposed (all)             & 0.7869           & -0.4\%          & 0.3152    & +0.2\%             & 0.7836           & +1.0\%          & 0.3206     & +2.1\% \\
\midrule
\multirow{7}{*}{\rotatebox{80}{\parbox{8em}{Qwen3-30B-A3B\\\relax[thinking mode]}}} & - & Baseline & 0.9075 [0.9378] & - [-] & 0.3904   & -                  & 0.8673 [0.9409]  & - [-]           & 0.4027     & -      \\
\cmidrule{2-11}
                                  & \multirow{2}{*}{2:4}  & Naïve top-k                & 0.8605 [0.9386]  & -4.7\%~~[+0.1\%] & 0.3590   & -3.1\%             & 0.8264 [0.9300]  & -4.1\% [-1.1\%] & 0.3831     & -2.0\% \\
                                  &                       & Proposed (l.s.)            & 0.9241 [0.9423]  & +1.7\% [+0.5\%]  & 0.3910   & +0.1\%             & 0.8848 [0.9417]  & +1.7\% [+0.1\%] & 0.3905     & -1.2\% \\
\cmidrule{2-11}
                                  & \multirow{2}{*}{4:8}  & Naïve top-k                & 0.9007 [0.9401]  & -0.7\%~~[+0.2\%] & 0.3747   & -1.6\%             & 0.8665 [0.9325]  & -0.1\% [-0.8\%] & 0.3739     & -2.9\% \\
                                  &                       & Proposed (l.s.)            & 0.9287 [0.9401]  & +2.1\% [+0.2\%]  & 0.3893   & -0.1\%             & 0.8741 [0.9420]  & +0.8\% [+0.1\%] & 0.3912     & -1.2\% \\
\cmidrule{2-11}
                                  & \multirow{2}{*}{8:16} & Naïve top-k                & 0.9181 [0.9279]  & +1.1\% [-1.0\%]~ & 0.3901   & -0.0\%             & 0.8620 [0.9401]  & -0.5\% [-0.1\%] & 0.3788     & -2.4\% \\
                                  &                       & Proposed (l.s.)            & 0.9287 [0.9401]  & +2.1\% [+0.2\%]  & 0.3888   & -0.2\%             & 0.8741 [0.9454]  & +0.8\% [+0.5\%] & 0.3958     & -0.7\% \\
\bottomrule
\end{tabular}
\label{tab3}
\end{table*}

Building on prior results, we further examine the effectiveness of integrating W8A8 quantization with the Amber Pruner algorithm. 
Standard post-training quantization (PTQ) is used: activations are quantized per tensor with SmoothQuant-based scaling, calibrated on 50 samples from the BoolQ dataset; weights are quantized per channel. Experiments show that setting the SmoothQuant hyperparameter to $\alpha$ = 0.10 amplifies activation outliers (see Figure \ref{fig5}), enhancing sparsity selectivity to improve pruning of low-importance channels.

The quantization strategies adopted for different models are summarized below:
\begin{itemize}
\item \textbf{LLaMA3.1-8B}: Skip quantization for all linear projections in the first 5 layers and all down\_proj layers.
\item \textbf{Qwen2-7B}: Skip all down\_proj layers.
\item \textbf{Qwen3-30B-A3B}: gate\_proj projections in all layers are excluded from quantization. A hybrid strategy is adopted, where attention modules use static W8A8 quantization, while MoE layers apply per-token dynamic quantization.
\end{itemize}

\subsubsection{Results}
Table \ref{tab2} summarizes the Zero-shot performance of Outstanding-sparse. The combined use of activation sparsification and weight quantization yields competitive results under consistent evaluation settings. Depending on the layer's quantization sensitivity, linear projections may be subject to quantization, pruning, or both. Key observations include:
\begin{itemize}
\item \textbf{Effectiveness of the Proposed Method}: The joint quantization and sparsification approach enables controllable accuracy loss. In some cases, Outstanding-sparse reshapes the activation distribution, exposing more sparsity and improving pruning effectiveness—a behavior not observed in the original Bfloat16 activations (see Figure \ref{fig5}).
\item \textbf{Sparsity as the Primary Accuracy Bottleneck}: While SQ-W8A8 serves as a lossless quantization baseline relative to Bfloat16, the majority of the accuracy degradation observed in Zero-shot evaluations is attributable to activation sparsification rather than quantization.
\item \textbf{Robustness on MoE Models}: MoE models show strong compatibility with Outstanding-sparse. Under default configurations, the 2:4 sparsity pattern results in only 1.5\% accuracy drop, while the 8:16 sparsity keeps the degradation below 1.0\%.
\end{itemize}

Table \ref{tab3} presents the generation performance of the proposed Amber Pruner and Outstanding-sparse. Both algorithms achieve strong performance on GSM8K \cite{cobbe2021gsm8k} and LongBench \cite{bai2023longbench}, with no significant accuracy degradation observed. Relevant examples are provided in \textbf{Appendix E}. Additionally, we report partial results for Qwen3-30B-A3B operating in \textbf{[thinking mode]}. This suggests that the impact on the KV cache from applying sparsity during the prefill phase is not substantial enough to compromise generation quality in the decoding stage. By confining sparsity to the prefill phase, the approach effectively alleviates computational overhead while maintaining output quality. Overall, it achieves a well-balanced trade-off between efficiency and performance.

In summary, our training-free algorithm Amber Pruner, applied during the prefill phase, achieves effective synergy with existing W8A8 quantization methods and is compatible with both Dense and MoE models. Experimental results show that the generation performance of sparse LLMs remains stable, with less than 1\% average accuracy loss observed in Zero-shot evaluations.

\section{Conclusion}
This work fully unleashes the potential of hardware-friendly N:M sparsity by introducing Amber Pruner—a training-free algorithm that imposes N:M activation sparsity during the prefill phase to alleviate computational bottlenecks. Unlike conventional weight sparsification, Amber Pruner capitalizes on the inherent sparsity of activations, achieving significant accuracy retention without relying on retraining or fine-tuning tricks. Its versatility is validated across diverse LLM architectures, with consistent performance retained. In particular, under the 8:16 sparsity pattern, Amber Pruner demonstrates strong potential for practical deployment. Although current hardware limitations, such as the lack of SpMM support, hinder observed acceleration gains, our findings pave the way for rethinking N:M sparsity as a viable and impactful direction for efficient LLM inference. We believe Amber Pruner marks an important step toward bridging algorithmic innovation and hardware optimization in next-generation AI systems.




\bibliography{aaai2026}

\end{document}


\maketitle

\section{Appendix A}

This section supplements Page 3 of the main paper. We conducted a comparative study between commonly used weight sparsification methods and a simple Naïve top-k activation sparsification approach, using LLaMA3.1-8B-Instruct \cite{grattafiori2024llama} as the base model. The evaluated weight sparsification methods include SparseGPT \cite{frantar2023sparsegpt}, Wanda \cite{sun2023simple}, and Pruner-Zero \cite{dong2024pruner}, while the activation sparsification method adopts a straightforward strategy: selecting the top-k activation values by magnitude to enforce N:M sparsity.

Sensitive layer skipping was not applied in this analysis. The benchmarking configuration is described in detail on Page 4 of the main text.

As illustrated in Table \ref{suppltab1}, although the weight sparsification methods adopt well-established academic techniques, the Naïve top-k activation sparsification consistently outperforms them. This indicates that even simple activation-based strategies can provide significant advantages over weight pruning approaches.

\section{Appendix B}
This section corresponds to Page 3 of the original text and primarily explains the theoretical derivation of the Wanda-like approach.

Inspired by the principles of \cite{frantar2023sparsegpt} and Wanda \cite{sun2023simple}, we derive an importance-based scoring function tailored for activation sparsity. Our goal is to identify which activation elements should be retained in order to minimize the perturbation to the output caused by sparsification.

Consider a linear projection layer with input activation vector $x \in \mathbb{R}^{d_{in}}$ and weight matrix $W \in \mathbb{R}^{d_{out} \times d_{in}}$. Let $x_m$ denote the sparsified version of $x$, where certain elements have been zeroed out. The change in the output can then be expressed as
\begin{equation}
\delta y = W \left ( x - x_m \right ) = - W m,
\end{equation}
where $m = x - x_m$ is the residual vector representing the pruned (zeroed) activation components. Assuming that the columns of $W$, i.e., $W_{:,j}$, are approximately orthogonal, which is a reasonable assumption in the context of LLMs, the output perturbation energy can be approximated as
\begin{equation}
{\left \| \delta y \right \|}_2^2 = {\left \| \sum_{j \in M}{x_j \cdot W_{:,j}}  \right \|}_2^2 \approx \sum_{j \in M}{x_j^2 \cdot {\left \| W_{:,j} \right \|}^2_2 },
\end{equation}
where $M$ is the index set of pruned activation elements. The approximation holds due to the near-orthogonality of the weight columns, which allows the cross-terms to be neglected in the approximation. Hence, the individual contribution of the $j$-th activation $x_j$ to the output error can be quantified as
\begin{equation}
\Delta_j = x_j^2 \cdot {\left \| W_{:,j} \right \|}_2^2.
\end{equation}
This suggests that to minimize total perturbation ${\left \| \delta y \right \|}_2^2$, we should preserve the activation elements with the highest contribution $\Delta_j$. Accordingly, we define an importance score for each activation as
\begin{equation}
S_j = \left | x_j \right | \cdot {\left \| W_{:,j} \right \|}_2.
\end{equation}
This score naturally balances the magnitude of the activation with the sensitivity of the corresponding output direction, as measured by the norm of the weight column.

However, during low-precision inference (e.g., INT8 or Bfloat16), activation values are susceptible to underflow when multiplied by very small norms. To enhance numerical stability, we normalize the weight norms within each channel by their minimum value
\begin{equation}
f\left ( W_{:,j} \right ) = \frac{{\left \| W_{:,j} \right \|}_2}{\min_{k} {\left \| W_{:,k} \right \|}_2}.
\end{equation}
The final normalized scoring function becomes
\begin{equation}
S_j = \left | x_j \right | \cdot f\left ( W_{:,j} \right ) = \left | x_j \right | \cdot \left ( \frac{{\left \| W_{:,j} \right \| }_2}{\min_{k}{\left \| W_{:,k} \right \| }_2}  \right ).
\end{equation}
This formulation matches the importance function proposed in the original Wanda paper, and in our context, provides a principled criterion for selecting activations to retain under an N:M structured sparsity constraint. 

\section{Appendix C}
As discussed on Page 3 of the main text, different layers or modules within an LLM often exhibit varying sensitivity to sparsification and quantization. To illustrate this, we present representative heatmaps from the LLaMA3.1-8B-Instruct model \cite{grattafiori2024llama}, where each heatmap visualizes the absolute activation magnitudes of different modules. In these visualizations, darker colors indicate higher absolute values of activation responses. Notably, black regions correspond to negative activations, while red regions represent positive activations.

From an intuitive perspective, the down projection module exhibits a wide dynamic range in its activation distribution. This makes it highly amenable to sparsification, as many values can be pruned without significantly impacting the output. However, its broad value distribution also implies low compatibility with quantization, where preserving numerical precision becomes more difficult.

In contrast, the output projection (o\_projection) often shows sharp, localized activations, frequently concentrated on a single token or a few positions. This pattern suggests that the o\_projection is less suitable for sparsification, since zeroing out even a small number of activations may lead to noticeable accuracy degradation.

These observations support the necessity of module-aware strategies in both sparsification and quantization pipelines for LLM acceleration.

\begin{figure}[htbp]
\centering
\includegraphics[width=\columnwidth]{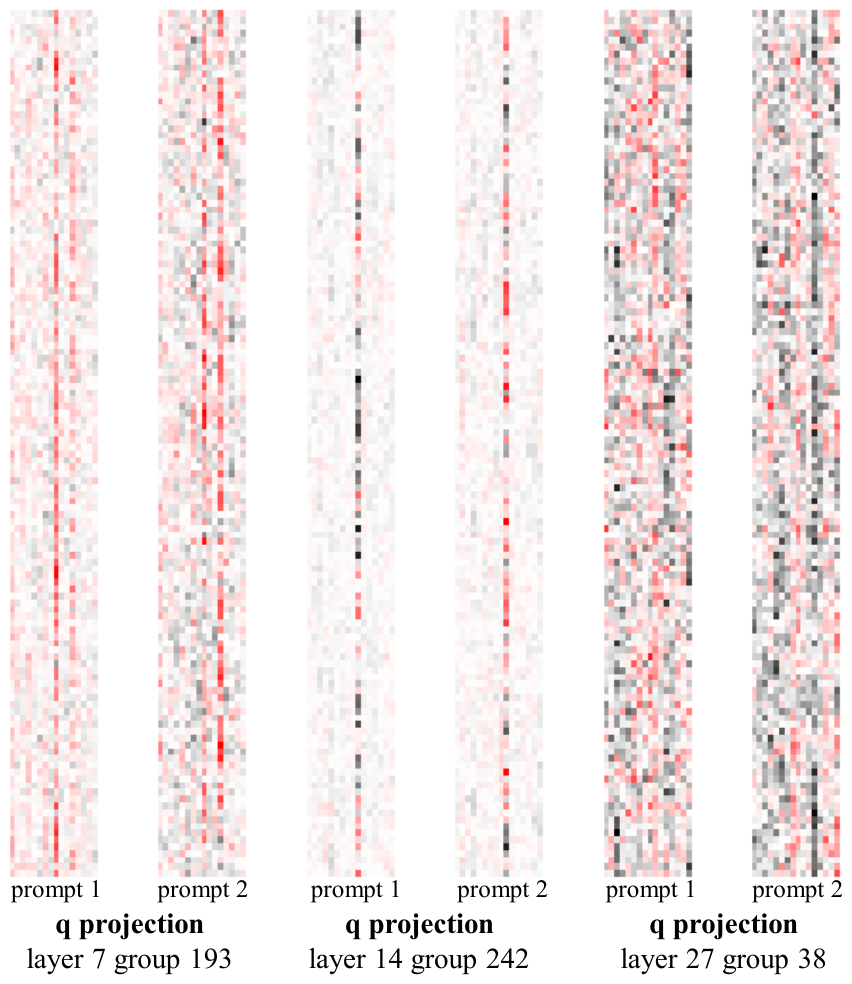}
\caption{Input activation heatmap of the q\_proj module.}
\label{supplFig1}
\end{figure}

\begin{figure}[htbp]
\centering
\includegraphics[width=\columnwidth]{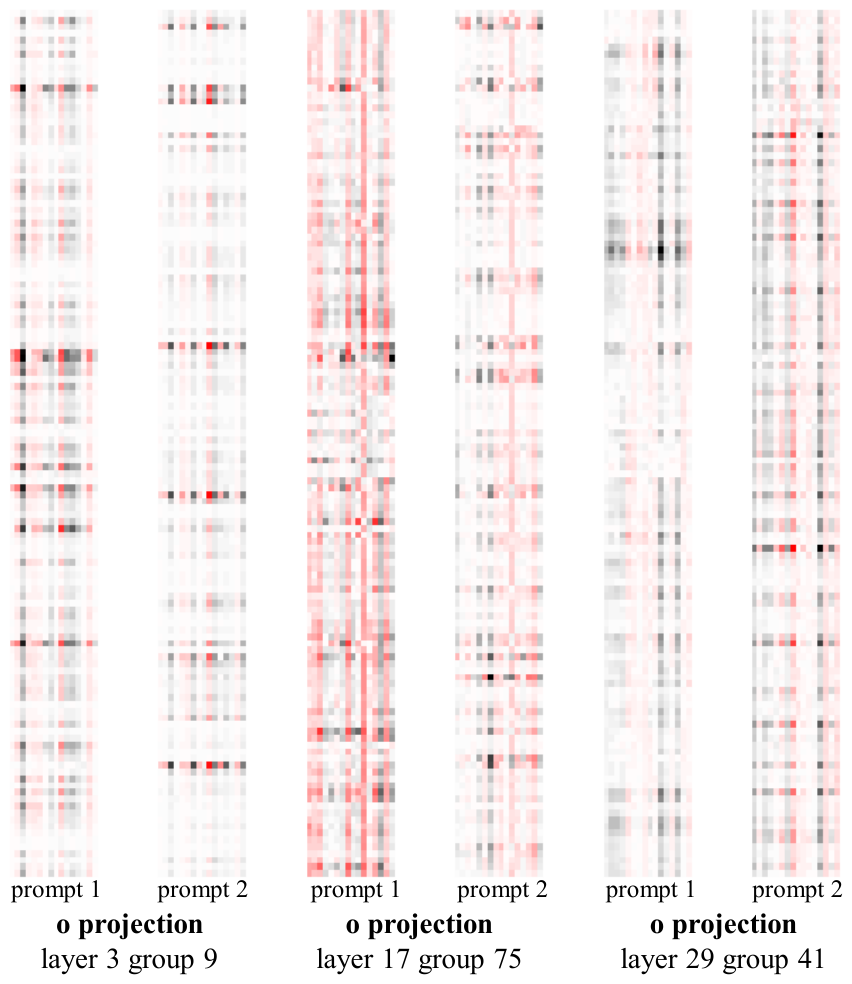}
\caption{Input activation heatmap of the o\_proj module.}
\label{supplFig2}
\end{figure}

\begin{figure}[htbp]
\centering
\includegraphics[width=\columnwidth]{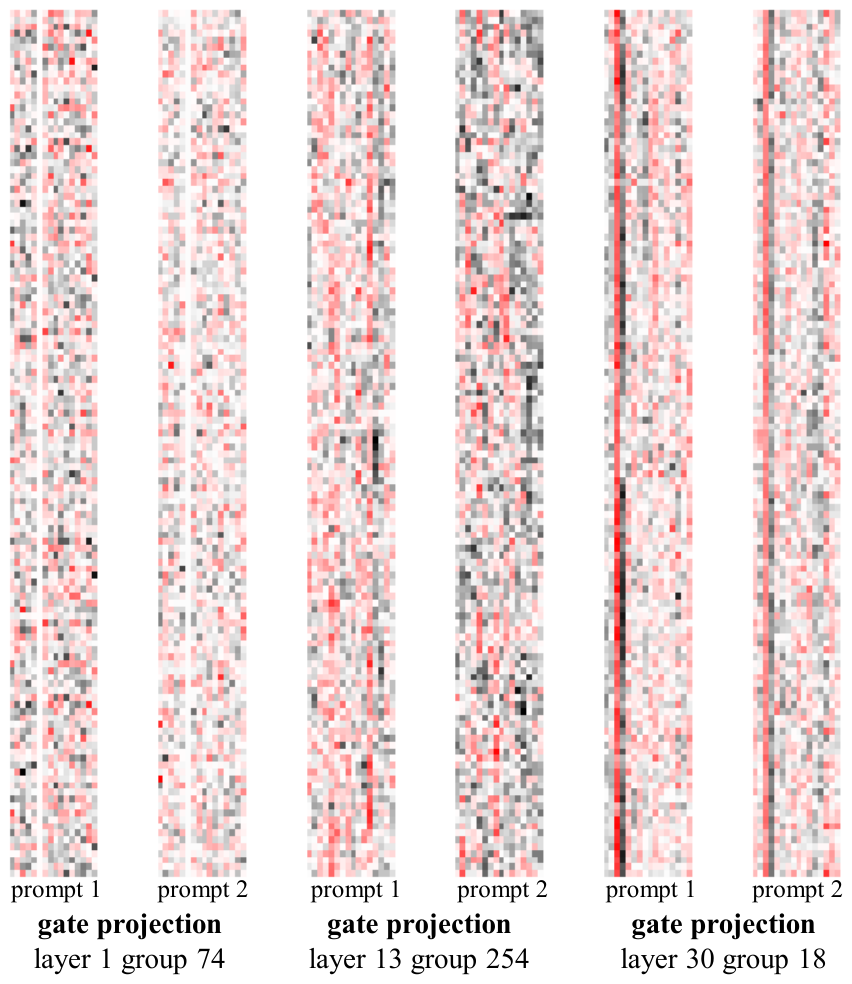}
\caption{Input activation heatmap of the gate\_proj module.}
\label{supplFig3}
\end{figure}

\begin{figure}[htbp]
\centering
\includegraphics[width=\columnwidth]{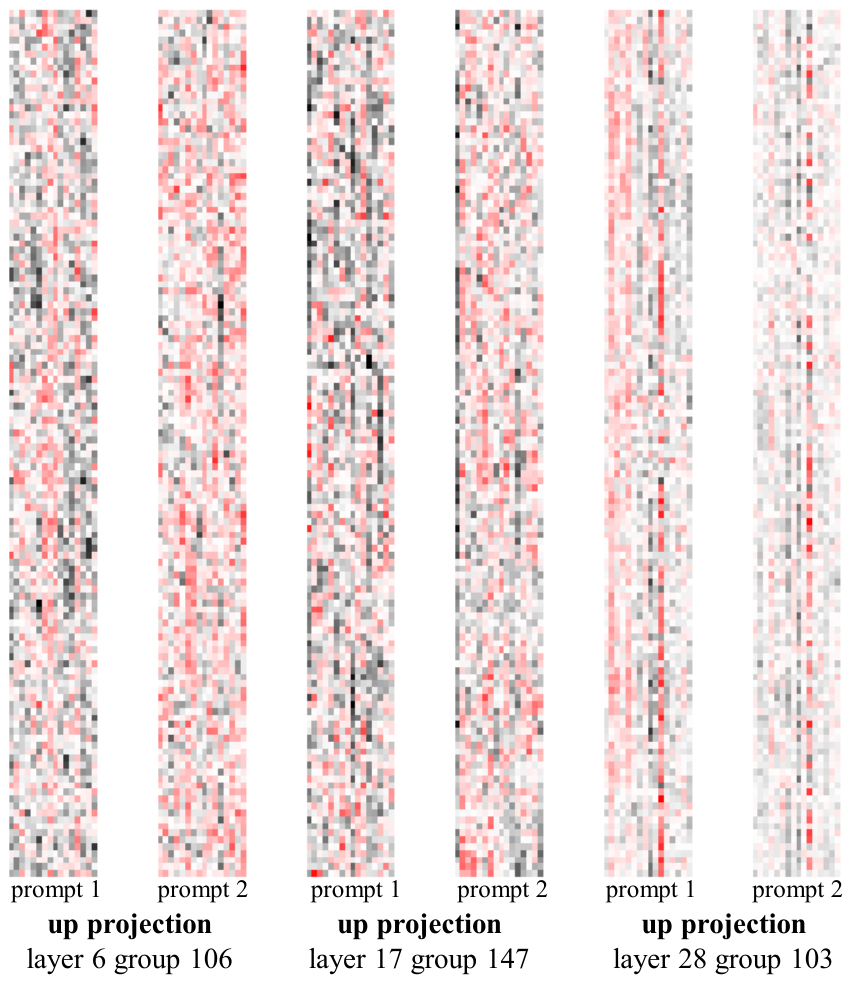}
\caption{Input activation heatmap of the up\_proj module.}
\label{supplFig4}
\end{figure}

\begin{figure}[htbp]
\centering
\includegraphics[width=\columnwidth]{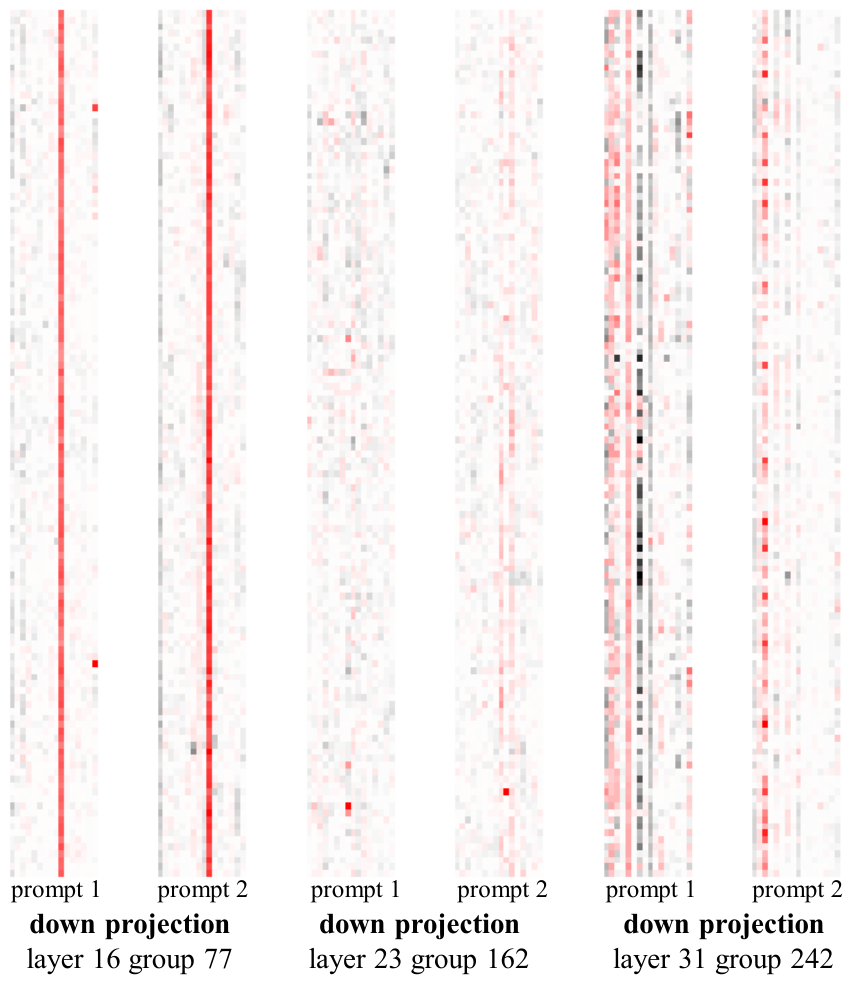}
\caption{Input activation heatmap of the down\_proj module.}
\label{supplFig5}
\end{figure}

\section{Appendix D}
\begin{figure}[h]
\centering
\includegraphics[width=0.95\columnwidth]{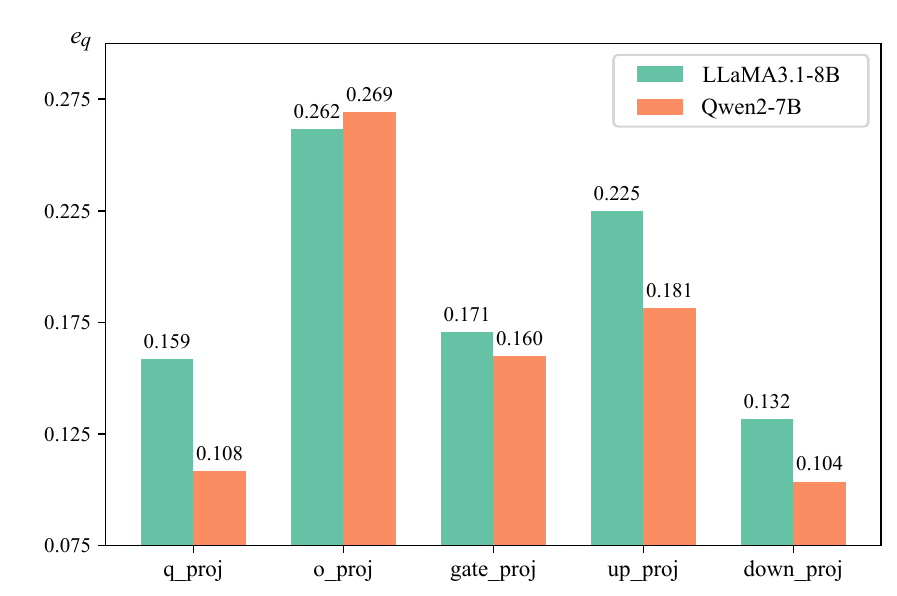}
\caption{Average sensitivity scores of linear projections.}
\label{supplFig6}
\end{figure}

This section supplements Page 4 of the main text by presenting average sensitivity scores for two representative models: LLaMA3.1-8B \cite{grattafiori2024llama} and Qwen2-7B \cite{team2024qwen2}. Each score $e_q$ represents the average sensitivity of a specific linear projection type across all layers. As shown in Figure \ref{supplFig6}, down\_proj exhibits the lowest average sensitivity, whereas o\_proj and up\_proj rank highest. In practice, sensitivity varies across layers, with those closer to the output generally displaying greater sensitivity, and thus warranting priority preservation during skipping.

\section{Appendix E}
This section provides representative examples of Amber Pruner and Outstanding-sparse on the GSM8K benchmark \cite{cobbe2021gsm8k}, serving as a supplement to Page 7 of the main text. GSM8K is a high-quality dataset specifically designed to assess a model’s problem-solving capabilities in grade-school math. Each question requires multi-step reasoning, basic arithmetic, and contextual understanding. Using LLaMA3.1-8B-Instruct \cite{grattafiori2024llama} as the base model, we illustrate the behavior of different sparsity strategies under varying compression levels. Correct outputs are highlighted in \textcolor{teal}{\textbf{green}}, while incorrect predictions are marked in \textcolor{red}{\textbf{red}}, providing intuitive insight into the model's robustness and limitations under structured sparsity.

\subsection{Assessing Amber Pruner on GSM8K}
\begin{mdframed}[linewidth=1.1pt]
\textbf{Input Context}: Question: Mitchell is making nachos for his family. He buys two bags of chips with 55 chips each. If his family has five members, how many chips does each person get if they all get the same number? Answer:
\vspace{0.5em}
\hrule
\vspace{0.5em}
\noindent\textbf{Bfloat16}: ...110 chips / 5 people = \textcolor{teal}{\textbf{22}} chips/person.\\
\noindent\textbf{8:16 Sparsity}: ...110 chips / 5 members = \textcolor{teal}{\textbf{22}} chips/member.\\
\noindent\textbf{4:8 Sparsity}: ...110 chips / 5 family members = \textcolor{teal}{\textbf{22}} chips per person.\\
\noindent\textbf{2:4 Sparsity}: ...110 chips / 5 people = \textcolor{teal}{\textbf{22}} chips/person.
\end{mdframed}

\begin{mdframed}[linewidth=1.1pt]
\textbf{Input Context}: Question: Raymond and Samantha are cousins. Raymond was born 6 years before Samantha. Raymond had a son at the age of 23. If Samantha is now 31, how many years ago was Raymond's son born? Answer:
\vspace{0.5em}
\hrule
\vspace{0.5em}
\noindent\textbf{Bfloat16}: ...Therefore, Raymond's son was born \textcolor{teal}{\textbf{14}} years ago.\\
\noindent\textbf{8:16 Sparsity}: ...So, \textcolor{teal}{\textbf{14}} years ago was Raymond's son born.\\
\noindent\textbf{4:8 Sparsity}: ...Therefore, Raymond's son was born \textcolor{teal}{\textbf{14}} years ago.\\
\noindent\textbf{2:4 Sparsity}: ...Therefore, Raymond's son was born \textcolor{red}{\textbf{2}} years ago.
\end{mdframed}

\begin{mdframed}[linewidth=1.1pt]
\textbf{Input Context}: Question: Nate's dog can dig six holes a day. He digs for 14 days while Nate is on vacation. ...How many weeks does it take him to fill in all the holes? Answer:
\vspace{0.5em}
\hrule
\vspace{0.5em}
\noindent\textbf{Bfloat16}: ...28 days / 7 days/week = 4 weeks. The answer is \textcolor{teal}{\textbf{4}}.\\
\noindent\textbf{8:16 Sparsity}: ...Therefore, it takes Nate \textcolor{teal}{\textbf{4}} weeks to fill in all the holes.\\
\noindent\textbf{4:8 Sparsity}: ...So, it will take Nate \textcolor{teal}{\textbf{4}} weeks to fill in all the holes.\\
\noindent\textbf{2:4 Sparsity}: ...So, it will take Nate approximately \textcolor{red}{\textbf{1.33}} weeks to fill in all the holes.
\end{mdframed}

\begin{mdframed}[linewidth=1.1pt]
\textbf{Input Context}: Question: Gloria is shoe shopping when she comes across a pair of boots that fit her shoe budget. ...If one pair of heels costs \$33 and the other costs twice as much, how many dollars are the boots? Answer:
\vspace{0.5em}
\hrule
\vspace{0.5em}
\noindent\textbf{Bfloat16}: ...the cost of the boots is 99 + 5 = \$104. The final answer is \textcolor{teal}{\textbf{\$104}}.\\
\noindent\textbf{8:16 Sparsity}: ...Add 5 to both sides: 104 = boots. The boots cost \textcolor{teal}{\textbf{\$104}}.\\
\noindent\textbf{4:8 Sparsity}: ...the boots cost 99 + 5 = 104 dollars. The final answer is \textcolor{teal}{\textbf{104}}.\\
\noindent\textbf{2:4 Sparsity}: ...Add 5 to both sides: 54.5 = boots. So, the boots cost \textcolor{red}{\textbf{\$54.50}}.
\end{mdframed}

\begin{mdframed}[linewidth=1.1pt]
\textbf{Input Context}: Question: Janet is collecting the results of Herbert Hoover Elementary School's annual standardized test. ...What is the school's overall pass rate? Answer:
\vspace{0.5em}
\hrule
\vspace{0.5em}
\noindent\textbf{Bfloat16}: ...(700 / 1000) * 100 = 70\%. So, the school's overall pass rate is \textcolor{teal}{\textbf{70\%}}.\\
\noindent\textbf{8:16 Sparsity}: ...So, the school's overall pass rate is \textcolor{teal}{\textbf{70\%}}.\\
\noindent\textbf{4:8 Sparsity}: ...So the overall pass rate is (460 / 1000) * 100 = 46\%. The answer is \textcolor{red}{\textbf{46}}.\\
\noindent\textbf{2:4 Sparsity}: ...overall pass rate is (460 / 1000) * 100 = 46\%. The answer is \textcolor{red}{\textbf{46}}.
\end{mdframed}

\begin{table*}[htbp]
\centering
\fontsize{8pt}{8pt}\selectfont
\caption{Comparison of weight sparsification and Naïve top-k activation sparsification on the LLaMA3.1-8B-Instruct model.}
\begin{tabular}{c|l|ccccccccc|c}
\toprule
\textbf{Rt.}             & \textbf{Methods}                  & \textbf{AC} & \textbf{AE} & \textbf{BQ} & \textbf{MMLU} & \textbf{OBQA} & \textbf{PIQA} & \textbf{RTE} & \textbf{WG} & \textbf{Avg.} & \textbf{Drop} \\
\midrule
-                        & Baseline: Bfloat16                & 0.5196      & 0.8178      & 0.8416      & 0.6807        & 0.3340        & 0.8003        & 0.6823       & 0.7411      & 0.6772        & -             \\ 
\midrule
\multirow{4}{*}{2:4}     & Activation Sparsity: Naïve top-k  & 0.3874      & 0.7184      & 0.7590      & 0.4919        & 0.2640        & 0.7285        & 0.6029       & 0.6417      & 0.5742        & -10.3\%       \\
                         & Weight Sparsity: SparseGPT        & 0.3515      & 0.6566      & 0.7318      & 0.4364        & 0.2160        & 0.7057        & 0.5957       & 0.6385      & 0.5415        & -13.6\%       \\
                         & Weight Sparsity: Wanda            & 0.2875      & 0.5947      & 0.6547      & 0.3618        & 0.1940        & 0.6828        & 0.5884       & 0.6125      & 0.4971        & -18.0\%       \\
                         & Weight Sparsity: Pruner-zero      & 0.2799      & 0.6077      & 0.6859      & 0.3264        & 0.1900        & 0.6882        & 0.5812       & 0.6022      & 0.4952        & -18.2\%       \\
\midrule
\multirow{4}{*}{4:8}     & Activation Sparsity: Naïve top-k  & 0.4249      & 0.7597      & 0.7939      & 0.5502        & 0.2720        & 0.7497        & 0.6101       & 0.6677      & 0.6035        & -7.4\%        \\
                         & Weight Sparsity: SparseGPT        & 0.3788      & 0.6970      & 0.7853      & 0.5056        & 0.2380        & 0.7410        & 0.5596       & 0.6859      & 0.5739        & -10.3\%       \\
                         & Weight Sparsity: Wanda            & 0.3464      & 0.6814      & 0.7502      & 0.4796        & 0.2240        & 0.7252        & 0.5776       & 0.6590      & 0.5554        & -12.2\%       \\
                         & Weight Sparsity: Pruner-zero      & 0.3379      & 0.6831      & 0.7153      & 0.4813        & 0.2380        & 0.7236        & 0.5668       & 0.6598      & 0.5507        & -12.7\%       \\
\bottomrule
\end{tabular}
\label{suppltab1}
\end{table*}

\newpage
\subsection{Assessing Outstanding-sparse on GSM8K}
\begin{mdframed}[linewidth=1.1pt]
\textbf{Input Context}: Janet’s ducks lay 16 eggs per day. ...She sells the remainder at the farmers' market daily for \$2 per fresh duck egg. How much in dollars does she make every day at the farmers' market? Answer:
\vspace{0.5em}
\hrule
\vspace{0.5em}
\noindent\textbf{Bfloat16}: ... she makes 9 * 2 = \textcolor{teal}{\textbf{\$18}} every day at the farmers' market.\\
\noindent\textbf{8:16 Sparsity}: ... she makes 9 x \$2 = \$\textless{}9*2=18\textgreater{}\textcolor{teal}{\textbf{18}} per day at the farmers' market.\\
\noindent\textbf{4:8 Sparsity}: ...So, she makes 9 * 2 = \textcolor{teal}{\textbf{\$18}} every day at the farmers' market.\\
\noindent\textbf{2:4 Sparsity}: ...9 * 2 = \$\textless{}9*2=18\textgreater{}18 every day at the farmers' market. The final answer is \textcolor{teal}{\textbf{\$18}}.
\end{mdframed}

\begin{mdframed}[linewidth=1.1pt]
\textbf{Input Context}: Bud makes homemade macaroni and cheese once a week. The pasta costs \$1.00 a box ... How much money does Bud spend on making macaroni and cheese in one year? Answer:
\vspace{0.5em}
\hrule
\vspace{0.5em}
\noindent\textbf{Bfloat16}: ...In one year, he spends 52 * \$10.00 = \$\textless{}52*10=520\textgreater{}\textcolor{teal}{\textbf{520}}.\\
\noindent\textbf{8:16 Sparsity}: ...The total yearly cost is \$10.00 * 52 = \$\textless{}10*52=520\textgreater{}\textcolor{teal}{\textbf{520}}.\\
\noindent\textbf{4:8 Sparsity}: ...the total cost per year is \$10.00 * 52 weeks = \$\textless{}10*52=520\textgreater{}\textcolor{teal}{\textbf{520}}.\\
\noindent\textbf{2:4 Sparsity}: ...making macaroni and cheese in one year is \$10 * 52 = \$\textless{}10*52=520\textgreater{}\textcolor{teal}{\textbf{520}}.
\end{mdframed}

\begin{mdframed}[linewidth=1.1pt]
\textbf{Input Context}: John and DeSean bought a bag of marshmallows to make s'mores together. ...how many S'mores can each kid have with the marshmallows left in the bag? Answer:
\vspace{0.5em}
\hrule
\vspace{0.5em}
\noindent\textbf{Bfloat16}: ...can have \textcolor{teal}{\textbf{7}} S'mores with the marshmallows left in the bag.\\
\noindent\textbf{8:16 Sparsity}: ...\textcolor{teal}{\textbf{7}} S'mores with the marshmallows left in the bag.\\
\noindent\textbf{4:8 Sparsity}: ...marshmallows they have: 14/2 = \textless{}14/2=7\textgreater{}\textcolor{teal}{\textbf{7}} S'mores.\\
\noindent\textbf{2:4 Sparsity}: ...number of marshmallows left in the bag: \textcolor{red}{\textbf{14}} S'mores.
\end{mdframed}

\begin{mdframed}[linewidth=1.1pt]
\textbf{Input Context}: Buford writes many checks every year. ...he writes a check to the pest and lawn service. How many checks does Buford write per year? Answer:
\vspace{0.5em}
\hrule
\vspace{0.5em}
\noindent\textbf{Bfloat16}: ...12 (electric) + 12 (gas) + 24 (church) + 4 (pest and lawn) = \textcolor{teal}{\textbf{52}} checks per year.\\
\noindent\textbf{8:16 Sparsity}: ...Therefore, Buford writes a total of 48 + 4 = \textcolor{teal}{\textbf{52}} checks per year.\\
\noindent\textbf{4:8 Sparsity}: ...checks/month * 12 months/year = \textcolor{red}{\textbf{48}} checks/year.\\
\noindent\textbf{2:4 Sparsity}: ...1 (gas bill) + 24 (church) + 48 (pest and lawn service) = \textcolor{red}{\textbf{74}} checks per year.
\end{mdframed}

\begin{mdframed}[linewidth=1.1pt]
\textbf{Input Context}: Walter is collecting money for charity. First he collects \$500 from his neighbors. Then he collects \$1500 from a fund he set up online ... How much is Walter's lawyer going to contribute? Answer:
\vspace{0.5em}
\hrule
\vspace{0.5em}
\noindent\textbf{Bfloat16}: ...donated a total of \$2200, Walter's lawyer will donate 3 x \$2200 = \$6600. The answer is \textcolor{teal}{\textbf{\$6600}}.\\
\noindent\textbf{8:16 Sparsity}: ...donate 3 x \$2200 = \$6600. The final answer is: \$\textless{}6600=6600\textgreater{}\textcolor{teal}{\textbf{6600}}.\\
\noindent\textbf{4:8 Sparsity}: ...which is 3 x \$700 = \$2100. The final answer is \textcolor{red}{\textbf{\$2100}}.\\
\noindent\textbf{2:4 Sparsity}: ...Therefore, Walter's lawyer is going to contribute \textcolor{red}{\textbf{\$6000}}.
\end{mdframed}
\bibliography{aaai2026}